\RequirePackage{fix-cm}
\documentclass[twocolumn]{svjour3}          
\smartqed  
\usepackage{graphicx}

\usepackage[numbers]{natbib}
\begin{document}\sloppy

\title{A microservice-based framework for exploring data selection in cross-building knowledge transfer
}


\author{Mouna Labiadh* \and 
Christian Obrecht \and 
Catarina Ferreira da Silva \and 
Parisa Ghodous} 


\institute{
M. Labiadh, P. Ghodous \at
LIRIS UMR5205, Univ Lyon, CNRS, Université Claude Bernard Lyon 1,F-69100, Villeurbanne, France \\
\email{\{mouna.labiadh*, parisa.ghodous\}@univ-lyon1.fr} 
\and
C. Obrecht \at
CETHIL UMR5008, Univ Lyon, CNRS, INSA-Lyon, F-69621, Villeurbanne, France \\
\email{christian.obrecht@insa-lyon.fr} 
\and
C. Ferreira da Silva \at
Instituto Universitário de Lisboa (ISCTE-IUL), Lisboa, Portugal \\
\email{catarina.ferreira.silva@iscte-iul.pt}
}
%

\date{Received: date / Accepted: date}

\maketitle

\begin{abstract}
Supervised deep learning has achieved remarkable success in various applications. Successful machine learning application however depends on the availability of sufficiently large amount of data. In the absence of data from the target domain, representative data collection from multiple sources is often needed. However, a model trained on existing multi-source data might generalize poorly on the unseen target domain. This problem is referred to as domain shift. In this paper, we explore the suitability of multi-source training data selection to tackle the domain shift challenge in the context of domain generalization. We also propose a microservice-oriented methodology for supporting this solution. We perform our experimental study on the use case of building energy consumption prediction. Experimental results suggest that minimal building description is capable of improving cross-building generalization performances when used to select energy consumption data.
\end{abstract}
\section{Introduction}

Predictive modeling in buildings plays an integral part in the efficient planning and operation of power systems. Adequate operational data are usually a prerequisite, especially when deep learning is adopted \cite{seyedzadeh2018machine,runge2019forecasting,li2017building}. Powerful machine learning models should rely on insightful utilization of relevant operational data in a sufficient amount. 


Nevertheless, building historical data are not always available, such as in newly built and renovated buildings \cite{filippidou2019effectiveness}. Renovation or replacement of existing buildings consider improving their energy efficiency based on energy saving measures (e.g. enhanced thermal insulation, highly energy-efficient electrical systems). It plays an important role in reducing the total energy consumption and lowering the greenhouse gas emissions of the existing building stock. Modeling of these buildings thus poses a challenge since that we do not have a priori knowledge about their improved energy consumption performance.

Already existing energy consumption data about other buildings can howbeit be obtained. The main idea of our work thus consists on leveraging representative data from multiple different (but related) source buildings. However, with possible domain shifts among multi-source and target data, it is improper to apply a single model via combining all multi-source data. Domain shift \cite{sugiyama2007mixture} is a key challenge where distributions mismatch across different data domains. Therefore, models trained on one or many source domains generalize poorly when applied to a different target domain. Namely, in our context, energy consumption profile in buildings depends considerably on several contextual factors, such as the building type (e.g. residential, commercial, office), size, age, location, etc. Combining energy data from disparate source buildings to model a target building from which no operational data are available, is consequently counterproductive and will adversely hurt the target performance.

Proposed approaches addressing the domain shift challenge are mainly classified into \textit{domain adaptation} and \textit{domain generalization}. Domain adaptation \cite{bousmalis2016domain,rozantsev2018beyond} utilizes labeled source data and unlabeled or sparsely labeled target data to obtain a well-performing model on the target domain. However, in several cases, the target data are not available. Domain Generalization (DG) \cite{blanchard2011generalizing,muandet2013domain} addresses such cases by utilizing multiple source domains. This paper considers the domain generalization area of research. We aim to train accurate predictive models that perform well on unseen target buildings which has no operational data, by leveraging knowledge from different but related source buildings. We also suppose to have a contextual description of the target building that can be utilized for source data selection. Data selection therefore enables to utilize most relevant source buildings based on their contextual similarity to the target building to be modeled.


For this purpose, we investigate the suitability of a data selection \cite{kouw2019review}  approach for cross-building domain generalization. To the best of our knowledge, our work is a first attempt to model a target building with minimal contextual information about it, and thus tackling the data unavailability problem by transferring knowledge from auxiliary buildings. Prior studies in this framework \cite{amarasinghe2017deep,ding2015personalized}  require labeled data of the building in question, such as historical consumption data, physical parameters of the building design, meteorological conditions, and/or information about the occupancy profiles, in order to train a reliable building energy consumption model. Our approach goes beyond state-of-the-art methods and proposes to transfer knowledge across multiple sources buildings while using minimal contextual information about the target building. This allows us to model buildings when we do not dispose of energy consumption data, such as in the case of renovated or newly-built buildings. To summarize, our main goal is to build a model that accurately predicts the future energy consumption of a previously unseen building, given training data from one or many selected buildings. For supporting our implementation, we propose a microservice-oriented system workflow that promotes scalability and elasticity when deployed in the cloud.

The remainder of this paper is structured as follows. Section~\ref{sec:1} presents a classification of domain generalization techniques. Section~\ref{sec:2} provides an overview on the microservices architecture of our proposed system and a definition of the predictive model we utilize. Section~\ref{sec:3} depicts the experimental setup and summarizes results. Section~\ref{sec:4} discusses experimental findings, and finally in Section~\ref{sec:5}, we draw conclusions and present an outlook and suggestions for future research. 

\section{Approaches to Domain Generalization}
\label{sec:1}

Domain generalization is a form of transfer learning, which applies expertise acquired in source domains to improve learning of different but related target domains \cite{pan2009survey}. Domain generalization focuses on the generalization ability of previously unseen target domains, in which no data are available during training. Proposed domain generalization approaches typically rely on the assumption that source domains and unseen target domains share common features that can be extracted. Hence, they seek to learn a domain agnostic representation or model. Domain generalization approaches proposed in literature may be roughly classified into three categories; (1) Data representation based techniques \cite{muandet2013domain,ghifary2015domain,khosla2012undoing,li2017deeper,li2018domain} that seek to learn domain agnostic representation that captures similarities across domains and where the domain discrepancy is minimized. (2) Ensembling techniques \cite{xu2014exploiting,bousmalis2016domain,ding2017deep,mancini2018best} that aim to build ensembles of per-domain models that will be then fused at test time. (3) Meta-learning based techniques \cite{li2018learning,balaji2018metareg} that rely on a model agnostic training procedure that trains any given model so that it mitigates domain shift between domains. 

Muandet et al. \cite{muandet2013domain} propose to learn new domain invariant feature representations by minimizing the dissimilarity across domains via domain-invariant component analysis and a kernel-based optimization algorithm. Ghifary et al. \cite{ghifary2015domain} propose a Multi-Task Auto-Encoder (MTAE) that extends auto-encoders into a model that jointly learns to perform self-domain data reconstruction and between-domain data reconstruction. Xu et al. \cite{xu2014exploiting} use learned low-rank exemplar-SVMs, which can be defined as a linear Support Vector Machine (SVM) classifier trained on a single positive training instance and all negative training instances, for both domain adaptation and domain generalization. For domain generalization, the authors propose to either equally fuse all exemplar classifiers, or use the exemplar classifiers in the latent domain which the target data more likely belongs to. Given multiple source datasets/domains, Khosla et al.\cite{khosla2012undoing} propose an SVM based approach, in which the learned weight vectors are common to all datasets. Li et al. \cite{li2017deeper} proposed a low-rank parameterized convolutional neural network model for end-to-end DG learning. Li et al. \cite{li2018learning} propose a Meta-Learning Domain Generalization (MLDG) approach. It consists in a model agnostic training procedure that can improve the domain generality of a base learner. This procedure is based on synthesizing virtual training and virtual testing domains within each mini-batch. The meta-optimization objective consists in minimizing the loss in the training domains, while simultaneously improving the loss in the testing domain.

Our work is more related to the model selection techniques. We borrow the per-domain model building idea described in \cite{xu2014exploiting}. However, we select domains rather than models and combine their respective data to form a representative training set. We assume in our case that we dispose of a minimal description of the target domain that will allow us to define our data selection criteria. Some examples of contextual descriptive features are building typology, area, year of construction, and number of occupants. 

Source domain selection has been proposed in the context of multi-source domain adaptation \cite{duan2012exploiting,chattopadhyay2012multisource}. This allows to select good source that are most relevant to the target domain and avoid negative transfer \cite{rosenstein2005transfer}. In \cite{duan2012exploiting}, authors proposed data-dependent regularizer for domain selection. Other works \cite{chattopadhyay2012multisource,duan2009domain} employed all source domains for adaptation but assigned different weights to different source domains. Weights are generally computed on the basis of some similarity measures between target and source domains. Several domain similarity metrics have been proposed for selection such as Kullback-Leibler divergence \cite{sugiyama2008direct}, Jensen-Shannon divergence, maximum mean discrepancy \cite{borgwardt2006integrating}, the Wasserstein metric \cite{shen2018wasserstein,xu2019wasserstein} or the Kolmogorov-Smirnoff statistic \cite{kouw2019review,huang2007correcting}. Even within one domain, adaptation performance varies significantly depending on the choice of data samples \cite{ruder2017data}. Other related work in the direction of data selection include using reinforcement learning to select data during neural network training \cite{fan2017learning}.

Both domain adaptation and domain generalization aim to learn an accurate model for the target domain by leveraging labeled data from the source domains. The difference between them is that for domain adaptation, unlabeled data and even a few labeled data from the target domain are utilized for adaptation. Whereas, for domain generalization, target data are not available. Our work falls within the latter case. We solely dispose of a minimal contextual description (metadata) to capture properties of the target domain for knowledge transfer. Some works have proposed to exploit available metadata about domains/tasks in addition to domain data to guide multi-domain learning and multi-task learning \cite{yang2014unified,yang2016multivariate}. Metadata in this work consisted of semantic descriptors of domain or task, and are combined with feature vectors during training. Rather than combining domain metadata and data, we are utilizing target domain metadata for source data selection. This way, we can address the domain generalization setting in which no target domain's data are available during training. We therefore propose in our context to select similar source buildings' data based on the target building's metadata and build a predictive model for the target building. The following section gives an in-depth description of our proposed methodology.

\section{The Proposed System}
\label{sec:2}

\begin{figure}[h]
    \centering
    \includegraphics[width=\columnwidth]{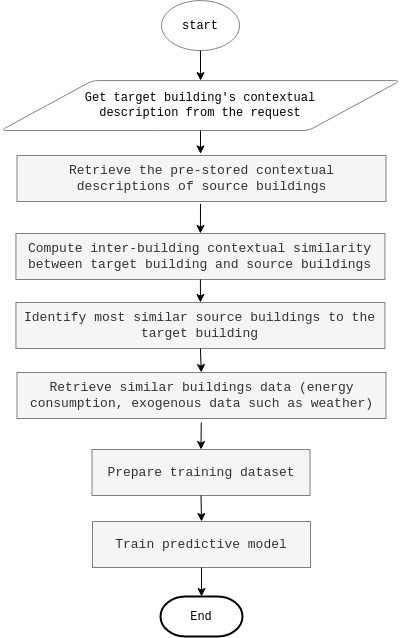}
    \caption{Flow chart describing the main steps performed by our system that provide cross-building knowledge transfer via source buildings selection. Rectangles show tasks. Parallelogram is used to show input data from the query.}
    \label{fig:flow}
\end{figure}

Our system main objective is to train an energy predictive model for an unseen target building based solely on its contextual description. In our special case, contextual descriptions concern high-level information about the target building we seek to model, e.g. typology, year of construction, location, etc. The training data of the target building's predictive model is obtained through an energy consumption data selection workflow. Data selection is performed based on the contextual similarity between the target building and the source buildings. The steps performed by our proposed system at each request are shown in Figure~\ref{fig:flow}.

Our approach consists in training a predictive model for an unseen target building via source data selection. Data selection is based on the similarity between the available source buildings and the unseen target building contextual descriptions. We assume that source buildings energy data and contextual descriptions are pre-collected and stored, whereas the target building contextual description is provided by system users. Once similar source buildings are identified, their corresponding energy data are retrieved. Energy data from buildings generally consist of historical energy consumption data along with critical exogenous variables such as weather conditions, holidays, etc. Retrieved source data from multiple sources are then combined to form a training dataset, and provided to the train a predictive model for the target building. A more detailed overview of our proposed workflow is provided in Figure~\ref{fig:workflow} of the following section.

Our system users are mainly building energy professionals and third-party building management systems which seek to accurately model a building on which operational energy data are not available. An accurate prediction of energy demands at the customer and building level will provide useful information to make decisions on energy generation and purchase. In this study, we attempt to explore the suitability of similar training data selection in the context of building energy consumption modeling.

\subsection{System Architecture and Data Specification}

We propose to establish a microservices-based architecture (MSA) for cross-building knowledge transfer. Each individual microservice is fully-independent, self-contained, and specific to a single task. Unlike monolithic applications, the MSA breaks down the application into a suite of flexible, independently deployable and loosely coupled modules that are accessible via a lightweight language-agnostic application programming interface (API). APIs are mainly based on asynchronous messaging protocols.

MSA offers several benefits, such as an increase in agility in development and delivery, resilience to failure, reliability in operation, maintainability, separation of concerns, and ease of deployment. Compared to service-oriented architecture (SOA), the core intent of the MSA pattern is to limit a service to a single purpose, enabling it to be fully decoupled and thus much more easily scaled and swapped out. Contrary to MSA, component sharing is one of the core tenets of SOA. SOA therefore relies on multiple services to fulfill a business request. Whereas MSA minimizes the need to share components through bounded context, which allows the coupling of a component and its data as a single unit with minimal dependencies. 

\begin{figure}[h]
    \centering
    \includegraphics[width=\columnwidth]{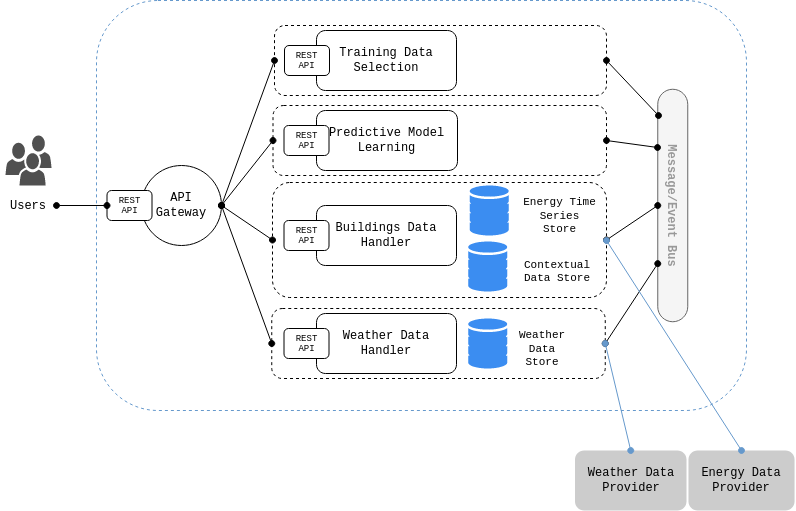}
    \caption{Our microservice architecture. Dotted rectangles represent individual microservices. Grey rectangles represent external third-party services.}
    \label{fig:sys}
\end{figure}

Figure~\ref{fig:sys} shows the various microservices and their coupling in our proposed system. Our system is capable of continuously ingesting and integrating data from external providers such as weather data and open energy data. Time series data about building energy consumption and weather data are respectively stored in the time series store and the weather data store. These two stores are linked together through the contextual information. In addition, contextual information provides a high-level description about the building environment, such as the year of construction, the building type, the size and the number of occupants.

The entry point of our system workflow is the data selection step. Via our system's API, users define the required use case by providing a key-value description of the unseen target building to model. No prior knowledge on the target building's energy consumption is needed. The most relevant time series data corresponding to most similar buildings, is then identified and selected. Similar buildings are identified based on the contextual information on the target building and the contextual information on other source buildings available within the system. The training data selection service loads contextual information from the contextual store via message queues.  

\begin{figure*}[h]
    \centering
    \includegraphics[width=\textwidth]{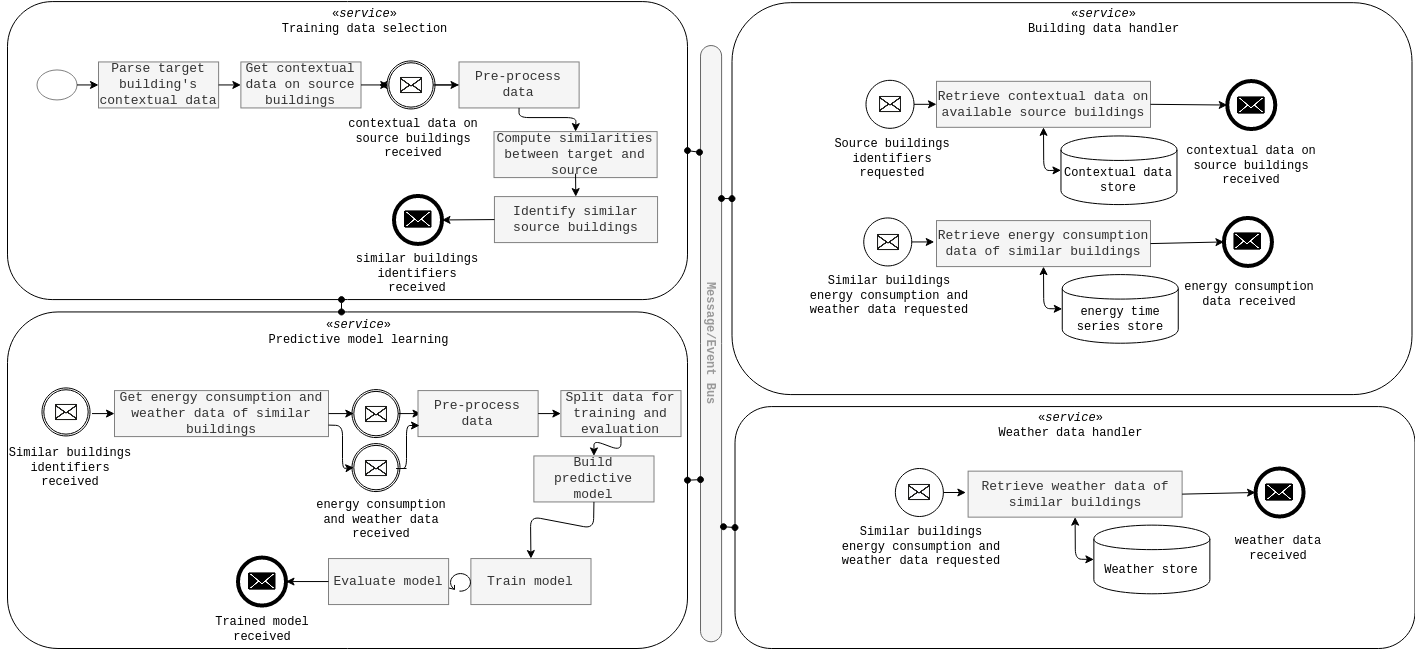}
    \caption{Overall representation of the microservices workflow and data flow.}
    \label{fig:workflow}
\end{figure*}

Once similar source buildings identifiers are available, predictive model learning service will load corresponding data from the time series store and/or weather data store via message queues. Training dataset will be then prepared using data transformation techniques, e.g. missing data imputation, outlier removal, etc. Finally, predictive learning model is trained in order to predict future energy consumption for a pre-defined forecasting horizon. In current work, we rely on a recurrent neural network for predictive modeling. The overall microservices workflow and data flow in our system is sketched in Figure~\ref{fig:workflow}.

Our training data selection workflow starts at each user request. It parses the contextual information about the target building contained in the request, and studies its similarity with pre-stored contextual information about available source buildings. Data in our system are shared between microservices following an event-based communication. Microservices therefore communicates via event messages. This enables loose coupling between collaborating microservices and privileges asynchronous behavior. Once similar source buildings are successfully identified, their identifiers are shared with the predictive model learning service. Building data and weather data handling microservices plays the role of data providers when selecting training data and training predictive models. Building data handler provides contextual information and energy consumption time series data about available source buildings to respectively the training data selection microservice and the predictive model learning microservice. Weather data handler provides exogenous weather data, such as air temperature, atmospheric pressure and wind speed, to the predictive model learning microservice.  

\begin{figure*}[h]
    \centering
    \includegraphics[width=.7\textwidth]{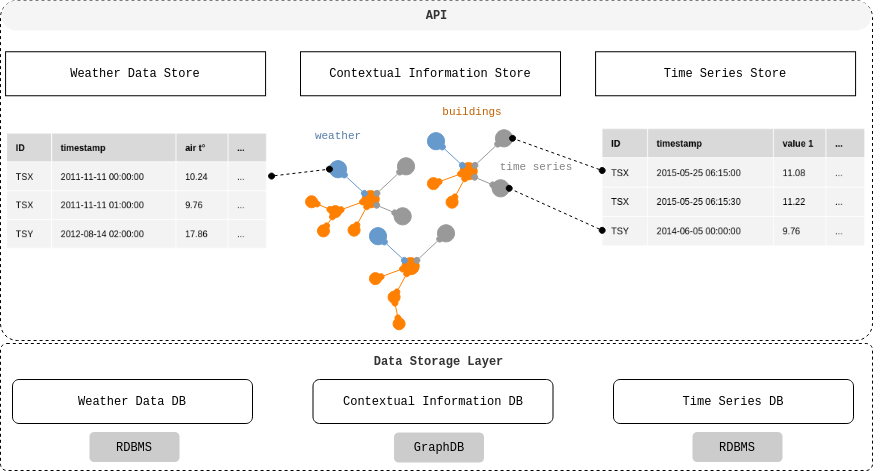}
    \caption{Contextual information and time series data management component diagram of our proposed system.}
    \label{fig:data}
\end{figure*}

To deal with potentially large-scale data, we rely on a multi-modal data store in the backend. Time series data are stored in a traditional relational data base management system (RDBMS). Our system is transparent to the specific database technology used. Contextual data about buildings and their associated time series is stored in a graph database. An overview on data management behind our API is shown in  Figure~\ref{fig:data}.

\subsection{Suitability of Training Data Selection}

In this study, we investigate the suitability of training data selection for cross-building knowledge transfer. The main logic behind our suitability study consists in training a predictive model using time series data of each building available in the dataset. Then, we test the cross-building generalization performance of each resulting predictive model, i.e. test it on other unseen buildings of the dataset. This will allow us to study the correlation between good generalization results between two buildings and similarity between their contextual information. We can therefore study the possibility to select representative building time series data based solely on available target building contextual information. 

Considering for example the task of energy consumption prediction for a residential building occupied by two people, built in 1990, renovated in 2014 and located in Lyon. Having no operational data about the target task, it is required to utilize other operational data on different source buildings to build a predictive model. However, different data collected from distant source buildings would necessarily induce negative transfer. We thus study a method that will enable us to select only similar buildings that will yield efficient cross-building prediction results. For example, we select residential buildings that are constructed around the same year, located in a region with similar climate, or subject to similar occupancy profile as the target building.


In our experimental study, we propose to compute similarities between target building and source buildings contextual information using a pairwise distance. Computational complexity of data selection is therefore O(n), where n is the total number of available source buildings. Predictive models then learn to predict future building-level aggregate energy consumption based on energy consumption history and both past and future climate data. In this work, we focus on the meteorological data factor by feeding our model with past and future climate data along with the aggregate past energy consumption. The motivation behind utilizing both future and past climate data are to attempt to capture the correlation between day-to-day weather conditions changes and the building's energy load profile. 

\subsection{Predictive Model Learning}

\begin{figure}[ht]
    \centering
    \includegraphics[width=.8\columnwidth]{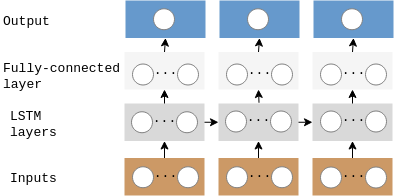}
    \caption{Architecture of the LSTM-RNN model.}
    \label{fig:modelarchi}
\end{figure}

Recently, deep learning is widely adopted for building energy consumption prediction tasks. Various deep learning model have been used, e.g. recurrent neural networks (RNN) \cite{kong2017short,wang2018short,kong2017short1}, sequence to sequence (Seq2Seq) models \cite{marino2016building}, combinations of convolutional neural network and recurrent neural network (CNN-RNN) \cite{tian2018deep,kim2019predicting}. In this work, we propose an unidirectional Long-Short Term Memory Recurrent Neural Network (LSTM-RNN) for the predictive modeling task. We present the architecture in Fig.~\ref{fig:modelarchi}. RNNs \cite{DBLP:journals/corr/Lipton15} are a powerful class of supervised machine learning models that are capable of modeling sequential data. They are artificial neural networks where connections between units can form cycles, which allows propagation of hidden state information from early parts of the sequence back to later point. LSTM \cite{hochreiter1997long} is a RNN architecture that helps to prevent the effect of vanishing and exploding gradients \cite{Pascanu:2013:DTR:3042817.3043083} often encountered in conventional recurrent networks. LSTM offers the ability to pass information selectively across sequence steps while processing sequential data one element at a time. 

Our model is trained to predict daily energy consumption of subsequent week. As input, we provide daily energy consumption of the previous week and climate time series of the subsequent week. 

Our training set $\mathcal{X} = \{(x^{(1)}, y^{(1)}), (x^{(2)}, x^{(2)}), ...\}$ is structured into time-based sequences of fixed length. Input sequences are denoted by $(x^{(1)}, x^{(2)}, ..., x^{(T)})$ where $T$ denotes the sequence length, and each value $x^{(t)} \in {\bf R}^7 \ {\rm for } t \in 1..T$. Feature vectors are composed of current week's aggregate energy consumption, air temperature, average horizontal solar irradiance,  wind speed, and these same features for subsequent week. Similarly, target sequences are denoted by $(y^{(1)}, y^{(2)}, ..., y^{(T)})$, where $y^{(t)} \in {\bf R}$ is a vector denoting the energy consumption at future time steps. The goal of the model is to predict future energy consumption $y^{(t)}$ from the input feature vector $x^{(t)}$.

The architecture of the network is composed of several hidden layers. It consists of one or more LSTM layers followed by one or more fully-connected layers. The output layer is a fully-connected layer with a linear activation function. The model is trained using the Root Mean Squared Error (RMSE). We also use the batch normalization mechanism \cite{ioffe2015batch} to address the internal covariate shift problem usually encountered in deep neural networks training. Training phase were conducted using Backpropagation Through Time (BPTT) \cite{werbos1990backpropagation} optimization algorithm in the context of LSTM networks. BPTT is commonly used to train recurrent networks. It “unfolds” the neural network in time by creating several copies of the recurrent units which can then be treated like a feed-forward network with tied weights. BPTT algorithm is known to be computationally efficient \cite{salehinejad2017recent, hochreiter1997long}, having a computational complexity per time step of O(W), where W is the number of weights.

During our experimental study, we explore variants of this architecture to fine-tune its hyperparameters, e.g. number of fully-connected layers, number of LSTM layers, etc. We retain the architecture variant that yields the best cross-domain and in-domain generalization results. 

\section{Experimental Setup}
\label{sec:3}

We perform our experimental studies on the use case of building energy consumption prediction. Our system transfers knowledge from several buildings, to one target building on which we assume we are facing a data unavailability problem.

\subsection{Dataset}

The proposed solution is experimentally evaluated using the REFIT Electrical Load Measurements dataset \cite{murray2017electrical}. The dataset contains cleaned electrical consumption measurements for 20 UK households at aggregate and appliance level. For each household, the whole house aggregate loads and nine individual appliance measurements at 8-second intervals were collected continuously over a period of approximately two years. During monitoring, the occupants were conducting their usual routines. In this paper, only the aggregate electrical consumption values for the whole house is used. We work with one-day resolution data which were obtained by summing the original data. 

\begin{figure}[ht]
    \centering
    \includegraphics[width=\columnwidth]{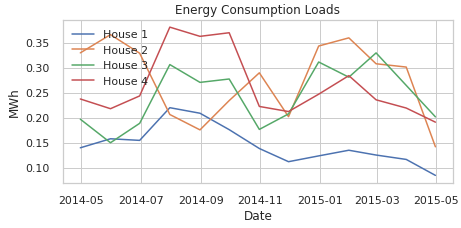}
    \caption{Monthly energy load profiles across buildings.}
    \label{fig:loads}
\end{figure}

In addition, climate data was also collected from a nearby weather station. Fig.~\ref{fig:loads} highlights the differences of energy load profiles across a subset of four buildings in the REFIT dataset. Descriptions about each building comprises information related to occupancy (number, age, gender, etc.), size, construction year, typology, and total number of owned appliances.

\begin{figure}[ht]
    \centering
    \includegraphics[width=3in]{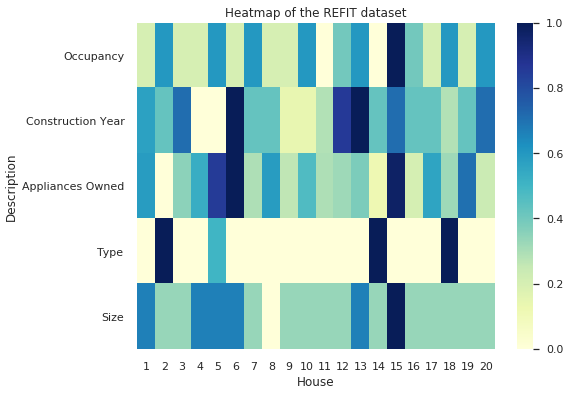}
    \caption{Heatmap of the REFIT dataset description after pre-processing; Missing data in one column were replaced with the most frequent value in that column, categorical values were label encoded, resulted values were scaled between 0 and 1.}
    \label{fig:refit_heat}
\end{figure}

In Fig.~\ref{fig:refit_heat}, we illustrate the REFIT dataset description with a heatmap. We consider five descriptive features for each building; the number of occupants, the construction year, the number of appliances, the building type, and the size. The number of occupants in the REFIT dataset varies from one and four occupants. The construction years of buildings are grouped into eight classes based on year intervals spanning from 1850 to post 2002. Three house types are present in the REFIT dataset; detached, semi-detached, and mid-terrace. Building sizes are computed based on number of bedrooms.

\begin{figure*}[ht]
    \centering
    \includegraphics[width=\textwidth]{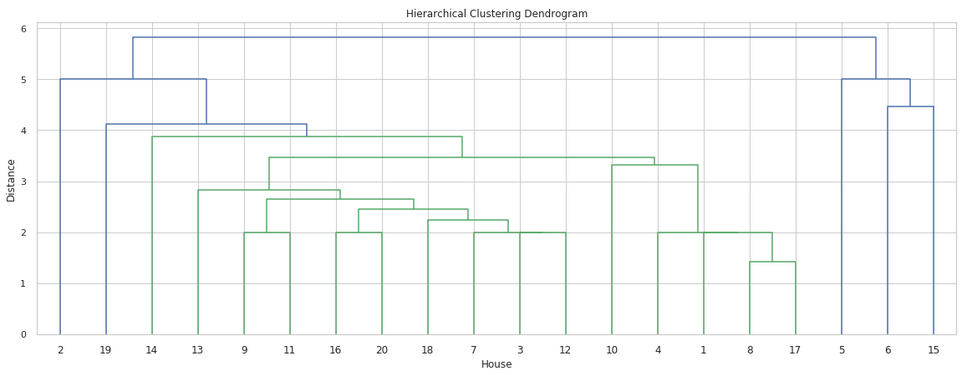}
    \caption{Dendrogram of the hierarchical clustering of REFIT households based on their descriptions. Clusters within which distance is below $70\%$ of the maximal cluster-wise distance Categorical are colored in green. features in the buildings' feature vectors were one-hot encoded. The distance used was the Euclidean distance.}
    \label{fig:refit_dendro}
\end{figure*}

To depict similarities between buildings, we start by hierarchically clustering them based on the provided description vectors. Categorical data was one-hot encoded as a further pre-processing step. We use the Euclidean distance to compute pair-wise similarities. Clustering results are illustrated in Fig.~\ref{fig:refit_dendro} by a dendrogram. The figure identifies a cluster of fourteen similar buildings, which is composed of the subset of the following buildings \{1, 3, 4, 7, 8, 9, 10, 11, 13, 14, 16, 17, 18, 20\}. Buildings 17 and 8 are identified as the most similar buildings in the dataset. Looking at their descriptions, they share the same number of occupants, building type, and construction year class. Building 17 also has only one more bedroom compared to building 8. Building pairs \{9, 11\}, and \{16, 20\} are also respectively identified as mutually similar.

\subsection{Model Training}

For each building, we use data between April 2014 and May 2015 for training. For cross-building evaluations, we use data between April 22nd, 2014 and June 1st, 2014. The whole dataset was scaled so all values will be between 0 and 1, using min-max normalization algorithm. The input and the output sequences are of length $7$. The input corresponds to a 7-dimensional feature vector. Our network is composed of two hidden layers; one LSTM layer of size $256$, and one fully-connected layer of size $128$. The Rectified Linear Unit (ReLU) is used as the non-linear activation function for hidden layers. The output layer consists of a fully-connected layer with linear activation function. The fine-tuning of weights is done using Gradient Descent algorithm with an exponentially decaying learning rate ranging between $10^{-3}$ and $10^{-5}$. Weights initialization follows a normal distribution with zero mean and standard deviation $\sigma=1$, whereas biases are initialized to zero. The gradients are back-propagated through timestep batches of length $80$. For the training epochs, we have fixed $1000$ as the maximum number. To avoid over-fitting, we have implemented an early stopping mechanism which breaks the training loop when training cost does not improve on the training set after $20$ epochs.

\subsection{Experimental Results}

Our goal is to achieve a good generalization performance by accurately predicting short-term energy consumption of unseen buildings. Therefore, we assess our proposed model using the Root Mean Squared Error (RMSE). RMSE is defined as the square root of the average squared distance between prediction and ground truth, using the formula:
\[
RMSE = \sqrt{\frac{1}{N} \sum_{i=1}^{N} (y_i - \hat{y}_i)^2} \,,
\]
where $y_i$ and $\hat{y}_i$ respectively denote the true value and the predicted value of the $i$-th data sample, and $N$ denotes the size of the dataset.

\begin{figure}[ht]
    \centering
    \includegraphics[width=\columnwidth]{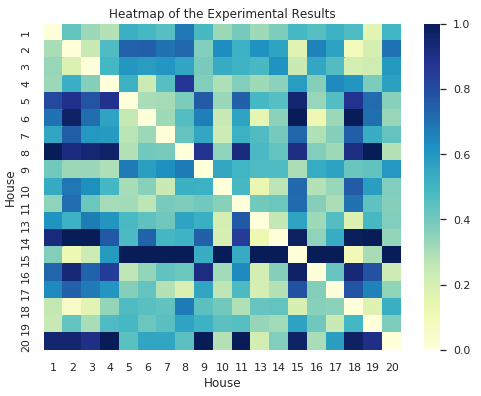}
    \caption{Heatmap of the experimental test errors; we trained 19 models, each of them on one single building. Each model was tested on each building. The y-axis represents buildings on which each model was trained, the x-axis represents the buildings on which each model was tested. The evaluation metric was RMSE. Final results were scaled between 0 and 1. House number $12$ was not considered due to insufficient training data.}
    \label{fig:results_heat}
\end{figure}

We trained $19$ models for each building following the same process. One building (number $12$) was not considered due to insufficient training data. Each model was tested on the remaining unseen buildings in order to study its cross-building transfer-ability. Fig.~\ref{fig:results_heat} depicts the predictions errors of cross-building model transfers as a heatmap. We can visually identify two clusters within each of them generalization performances are high. These clusters are respectively composed of the following subsets of buildings \{2, 3, 18, 19\} and \{5, 6, 7\}. We also notice that buildings 13 and 14 are mutually similar and that models trained on buildings 10 and 17 generalize well when applied to them during inference mode. Furthermore, we can visually conclude that all trained models perform poorly when applied to building 15. Model trained on building 15 also has poor generalization performances when applied to the remaining unseen buildings.

\begin{figure*}[ht]
    \centering
    \includegraphics[width=\textwidth]{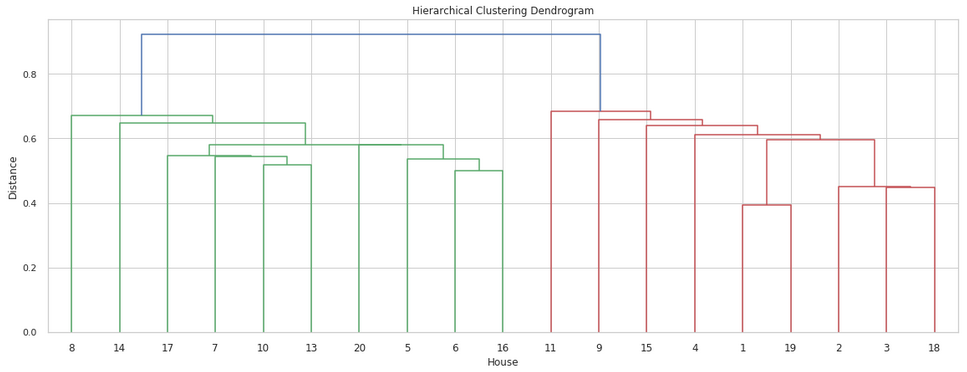}
    \caption{Dendrogram of the hierarchical clustering of REFIT households based on experimental cross-building prediction results. Clusters within which distance is below $70\%$ of the maximal cluster-wise distance Categorical are colored in green and red. The distance used was the Euclidean distance.}
    \label{fig:results_dendro}
\end{figure*}

We then seek to examine similar buildings based on these results; our assumption is that similar buildings models are transferable among each other. Hence, a model that is trained on a building $i$ will generalize well when applied to a building $j$ if buildings $i$ and $j$ are similar. We start by processing the experimental results matrix (Fig.~\ref{fig:results_heat}) to transform it to a distance matrix. For this purpose, we simply compute pairwise averages between each element at row $i$ and column $j$ and its corresponding element at row $j$ and column $i$. Drawn clusters from this distance matrix are illustrated in Fig.~\ref{fig:results_dendro} using a dendrogram. We use the Euclidean distance to compute pair-wise similarities. Fig.~\ref{fig:results_dendro} identifies two main clusters, which are respectively composed of the following subsets of buildings \{5, 6, 7, 8, 10, 13, 14, 16, 17, 20, 16\} and \{1, 2, 3, 4, 9, 11, 15, 18, 19\}.

\section{Discussion}
\label{sec:4}

From Fig.~\ref{fig:refit_dendro} and Fig.~\ref{fig:results_dendro}, we can notice that buildings 8 and 17 which were the most similar based on their descriptions are clustered under the same cluster based on their cross-domain generalization errors. This means that models trained on building 8 will generalize well when applied to building 17 during inference mode, and vice versa. Similarly, the two sets of buildings \{9, 11\}, and \{16, 20\} are identified as similar in both clustering schemes; based on descriptions and cross-domain generalization errors. Furthermore, poor cross-domain generalization performances of building 15 (Fig.~\ref{fig:results_heat}) is explainable by its dissimilarity with the rest of buildings (Fig.~\ref{fig:refit_dendro}).

We may therefore suggest that buildings, that are judged similar based solely on their descriptions, do yield to good prediction results when performing cross-building knowledge transfer.

In the context of this study, we have leveraged a very restricted set of building descriptions, i.e. number of occupants, typology, size, etc. Therefore, we believe that more heterogeneous and broader building descriptions (e.g. different types and locations) would help to select similar data more accurately and more reliably, and would make results more consistent. Furthermore, and due to the large variety of building typologies and design, and uncertainties surrounding its environment and occupancy patterns, we consider that data selection approaches based on similarity metrics are essential in order to perform large-scale and accurate cross-domain domain generalization.   

\section{Conclusion and Perspectives}
\label{sec:5}

This paper discusses the suitability of the data selection approach for cross-building knowledge transfer. Evaluation work was conducted on the case study of building energy consumption modeling. For this purpose, we have trained per-building models and studied their transferability across other unseen buildings. Experimental results show that minimal building descriptions are capable of guiding domain generalization applications in the context of energy modeling, by identifying similar buildings. Overall, we believe our results confirm the suitability of data selection mechanisms that are based on similarities of building minimal descriptions.

We also propose a microservice-oriented architecture that offers increased evolvability and scalability of the system as well as accelerated development velocity. 

Future  work  involves  exploring  and  reporting  the behavior of our approach with more larger scale and higher heterogeneity data sets. We also intend to extend our system by automating the data selection algorithm based on user queries. User queries will contain the description of the target building to which we want to transfer knowledge.

%
%
%
%
 \bibliographystyle{spbasic} 
 \bibliography{mybibliography}

\begin{thebibliography}{48}
\providecommand{\natexlab}[1]{#1}
\providecommand{\url}[1]{{#1}}
\providecommand{\urlprefix}{URL }
\expandafter\ifx\csname urlstyle\endcsname\relax
  \providecommand{\doi}[1]{DOI~\discretionary{}{}{}#1}\else
  \providecommand{\doi}{DOI~\discretionary{}{}{}\begingroup
  \urlstyle{rm}\Url}\fi
\providecommand{\eprint}[2][]{\url{#2}}

\bibitem[{Amarasinghe et~al.(2017)Amarasinghe, Marino, and
  Manic}]{amarasinghe2017deep}
Amarasinghe K, Marino DL, Manic M (2017) Deep neural networks for energy load
  forecasting. In: 2017 IEEE 26th International Symposium on Industrial
  Electronics (ISIE), pp 1483--1488

\bibitem[{Balaji et~al.(2018)Balaji, Sankaranarayanan, and
  Chellappa}]{balaji2018metareg}
Balaji Y, Sankaranarayanan S, Chellappa R (2018) Metareg: Towards domain
  generalization using meta-regularization. In: Advances in Neural Information
  Processing Systems, pp 998--1008

\bibitem[{Blanchard et~al.(2011)Blanchard, Lee, and
  Scott}]{blanchard2011generalizing}
Blanchard G, Lee G, Scott C (2011) Generalizing from several related
  classification tasks to a new unlabeled sample. In: Advances in neural
  information processing systems, pp 2178--2186

\bibitem[{Borgwardt et~al.(2006)Borgwardt, Gretton, Rasch, Kriegel,
  Sch{\"o}lkopf, and Smola}]{borgwardt2006integrating}
Borgwardt KM, Gretton A, Rasch MJ, Kriegel HP, Sch{\"o}lkopf B, Smola AJ (2006)
  Integrating structured biological data by kernel maximum mean discrepancy.
  Bioinformatics 22(14):e49--e57

\bibitem[{Bousmalis et~al.(2016)Bousmalis, Trigeorgis, Silberman, Krishnan, and
  Erhan}]{bousmalis2016domain}
Bousmalis K, Trigeorgis G, Silberman N, Krishnan D, Erhan D (2016) Domain
  separation networks. In: Advances in Neural Information Processing Systems,
  pp 343--351

\bibitem[{Chattopadhyay et~al.(2012)Chattopadhyay, Sun, Fan, Davidson,
  Panchanathan, and Ye}]{chattopadhyay2012multisource}
Chattopadhyay R, Sun Q, Fan W, Davidson I, Panchanathan S, Ye J (2012)
  Multisource domain adaptation and its application to early detection of
  fatigue. ACM Transactions on Knowledge Discovery from Data (TKDD) 6(4):1--26

\bibitem[{Ding et~al.(2015)Ding, Neumann, Stamm, Beigl, Inoue, and
  Pan}]{ding2015personalized}
Ding Y, Neumann MA, Stamm E, Beigl M, Inoue S, Pan X (2015) A personalized load
  forecasting enhanced by activity information. In: 2015 IEEE First
  International Smart Cities Conference (ISC2), pp 1--6

\bibitem[{Ding and Fu(2017)}]{ding2017deep}
Ding Z, Fu Y (2017) Deep domain generalization with structured low-rank
  constraint. IEEE Transactions on Image Processing 27(1):304--313

\bibitem[{Duan et~al.(2009)Duan, Tsang, Xu, and Chua}]{duan2009domain}
Duan L, Tsang IW, Xu D, Chua TS (2009) Domain adaptation from multiple sources
  via auxiliary classifiers. In: Proceedings of the 26th Annual International
  Conference on Machine Learning, pp 289--296

\bibitem[{Duan et~al.(2012)Duan, Xu, and Chang}]{duan2012exploiting}
Duan L, Xu D, Chang SF (2012) Exploiting web images for event recognition in
  consumer videos: A multiple source domain adaptation approach. In: 2012 IEEE
  Conference on Computer Vision and Pattern Recognition, IEEE, pp 1338--1345

\bibitem[{Fan et~al.(2017)Fan, Tian, Qin, Bian, and Liu}]{fan2017learning}
Fan Y, Tian F, Qin T, Bian J, Liu TY (2017) Learning what data to learn. arXiv
  preprint arXiv:170208635

\bibitem[{Filippidou et~al.(2019)Filippidou, Nieboer, and
  Visscher}]{filippidou2019effectiveness}
Filippidou F, Nieboer N, Visscher H (2019) Effectiveness of energy renovations:
  a reassessment based on actual consumption savings. Energy Efficiency
  12(1):19--35

\bibitem[{Ghifary et~al.(2015)Ghifary, Bastiaan~Kleijn, Zhang, and
  Balduzzi}]{ghifary2015domain}
Ghifary M, Bastiaan~Kleijn W, Zhang M, Balduzzi D (2015) Domain generalization
  for object recognition with multi-task autoencoders. In: Proceedings of the
  IEEE international conference on computer vision, pp 2551--2559

\bibitem[{Hochreiter and Schmidhuber(1997)}]{hochreiter1997long}
Hochreiter S, Schmidhuber J (1997) Long short-term memory. Neural computation
  9(8):1735--1780

\bibitem[{Huang et~al.(2007)Huang, Gretton, Borgwardt, Sch{\"o}lkopf, and
  Smola}]{huang2007correcting}
Huang J, Gretton A, Borgwardt K, Sch{\"o}lkopf B, Smola AJ (2007) Correcting
  sample selection bias by unlabeled data. In: Advances in neural information
  processing systems, pp 601--608

\bibitem[{Ioffe and Szegedy(2015)}]{ioffe2015batch}
Ioffe S, Szegedy C (2015) Batch normalization: Accelerating deep network
  training by reducing internal covariate shift. arXiv preprint arXiv:150203167

\bibitem[{Khosla et~al.(2012)Khosla, Zhou, Malisiewicz, Efros, and
  Torralba}]{khosla2012undoing}
Khosla A, Zhou T, Malisiewicz T, Efros AA, Torralba A (2012) Undoing the damage
  of dataset bias. In: European Conference on Computer Vision, Springer, pp
  158--171

\bibitem[{Kim and Cho(2019)}]{kim2019predicting}
Kim TY, Cho SB (2019) Predicting residential energy consumption using cnn-lstm
  neural networks. Energy 182:72--81

\bibitem[{Kong et~al.(2017{\natexlab{a}})Kong, Dong, Hill, Luo, and
  Xu}]{kong2017short}
Kong W, Dong ZY, Hill DJ, Luo F, Xu Y (2017{\natexlab{a}}) Short-term
  residential load forecasting based on resident behaviour learning. IEEE
  Transactions on Power Systems 33(1):1087--1088

\bibitem[{Kong et~al.(2017{\natexlab{b}})Kong, Dong, Jia, Hill, Xu, and
  Zhang}]{kong2017short1}
Kong W, Dong ZY, Jia Y, Hill DJ, Xu Y, Zhang Y (2017{\natexlab{b}}) Short-term
  residential load forecasting based on lstm recurrent neural network. IEEE
  Transactions on Smart Grid 10(1):841--851

\bibitem[{Kouw and Loog(2019)}]{kouw2019review}
Kouw WM, Loog M (2019) A review of domain adaptation without target labels.
  IEEE transactions on pattern analysis and machine intelligence

\bibitem[{Li et~al.(2017{\natexlab{a}})Li, Ding, Zhao, Yi, and
  Zhang}]{li2017building}
Li C, Ding Z, Zhao D, Yi J, Zhang G (2017{\natexlab{a}}) Building energy
  consumption prediction: An extreme deep learning approach. Energies
  10(10):1525

\bibitem[{Li et~al.(2017{\natexlab{b}})Li, Yang, Song, and
  Hospedales}]{li2017deeper}
Li D, Yang Y, Song YZ, Hospedales TM (2017{\natexlab{b}}) Deeper, broader and
  artier domain generalization. In: Proceedings of the IEEE International
  Conference on Computer Vision, pp 5542--5550

\bibitem[{Li et~al.(2018{\natexlab{a}})Li, Yang, Song, and
  Hospedales}]{li2018learning}
Li D, Yang Y, Song YZ, Hospedales TM (2018{\natexlab{a}}) Learning to
  generalize: Meta-learning for domain generalization. In: Thirty-Second AAAI
  Conference on Artificial Intelligence

\bibitem[{Li et~al.(2018{\natexlab{b}})Li, Jialin~Pan, Wang, and
  Kot}]{li2018domain}
Li H, Jialin~Pan S, Wang S, Kot AC (2018{\natexlab{b}}) Domain generalization
  with adversarial feature learning. In: Proceedings of the IEEE Conference on
  Computer Vision and Pattern Recognition, pp 5400--5409

\bibitem[{Lipton(2015)}]{DBLP:journals/corr/Lipton15}
Lipton ZC (2015) A critical review of recurrent neural networks for sequence
  learning. CoRR abs/1506.00019,
  \urlprefix\url{http://arxiv.org/abs/1506.00019}

\bibitem[{Mancini et~al.(2018)Mancini, Bul{\`o}, Caputo, and
  Ricci}]{mancini2018best}
Mancini M, Bul{\`o} SR, Caputo B, Ricci E (2018) Best sources forward: domain
  generalization through source-specific nets. In: 2018 25th IEEE International
  Conference on Image Processing (ICIP), pp 1353--1357

\bibitem[{Marino et~al.(2016)Marino, Amarasinghe, and
  Manic}]{marino2016building}
Marino DL, Amarasinghe K, Manic M (2016) Building energy load forecasting using
  deep neural networks. In: IECON 2016-42nd Annual Conference of the IEEE
  Industrial Electronics Society, IEEE, pp 7046--7051

\bibitem[{Muandet et~al.(2013)Muandet, Balduzzi, and
  Sch{\"o}lkopf}]{muandet2013domain}
Muandet K, Balduzzi D, Sch{\"o}lkopf B (2013) Domain generalization via
  invariant feature representation. In: International Conference on Machine
  Learning, pp 10--18

\bibitem[{Murray et~al.(2017)Murray, Stankovic, and
  Stankovic}]{murray2017electrical}
Murray D, Stankovic L, Stankovic V (2017) An electrical load measurements
  dataset of united kingdom households from a two-year longitudinal study.
  Scientific data 4:160122

\bibitem[{Pan and Yang(2009)}]{pan2009survey}
Pan SJ, Yang Q (2009) A survey on transfer learning. IEEE Transactions on
  knowledge and data engineering 22(10):1345--1359

\bibitem[{Pascanu et~al.(2013)Pascanu, Mikolov, and
  Bengio}]{Pascanu:2013:DTR:3042817.3043083}
Pascanu R, Mikolov T, Bengio Y (2013) On the difficulty of training recurrent
  neural networks. In: Proceedings of the 30th International Conference on
  International Conference on Machine Learning - Volume 28, JMLR.org, ICML'13,
  pp III--1310--III--1318,
  \urlprefix\url{http://dl.acm.org/citation.cfm?id=3042817.3043083}

\bibitem[{Rosenstein et~al.(2005)Rosenstein, Marx, Kaelbling, and
  Dietterich}]{rosenstein2005transfer}
Rosenstein MT, Marx Z, Kaelbling LP, Dietterich TG (2005) To transfer or not to
  transfer. In: NIPS 2005 workshop on transfer learning, vol 898, pp 1--4

\bibitem[{Rozantsev et~al.(2018)Rozantsev, Salzmann, and
  Fua}]{rozantsev2018beyond}
Rozantsev A, Salzmann M, Fua P (2018) Beyond sharing weights for deep domain
  adaptation. IEEE Transactions on Pattern Analysis and Machine Intelligence

\bibitem[{Ruder et~al.(2017)Ruder, Ghaffari, and Breslin}]{ruder2017data}
Ruder S, Ghaffari P, Breslin JG (2017) Data selection strategies for
  multi-domain sentiment analysis. arXiv preprint arXiv:170202426

\bibitem[{Runge and Zmeureanu(2019)}]{runge2019forecasting}
Runge J, Zmeureanu R (2019) Forecasting energy use in buildings using
  artificial neural networks: a review. Energies 12(17):3254

\bibitem[{Salehinejad et~al.(2017)Salehinejad, Sankar, Barfett, Colak, and
  Valaee}]{salehinejad2017recent}
Salehinejad H, Sankar S, Barfett J, Colak E, Valaee S (2017) Recent advances in
  recurrent neural networks. arXiv preprint arXiv:180101078

\bibitem[{Seyedzadeh et~al.(2018)Seyedzadeh, Rahimian, Glesk, and
  Roper}]{seyedzadeh2018machine}
Seyedzadeh S, Rahimian FP, Glesk I, Roper M (2018) Machine learning for
  estimation of building energy consumption and performance: a review.
  Visualization in Engineering 6(1):5

\bibitem[{Shen et~al.(2018)Shen, Qu, Zhang, and Yu}]{shen2018wasserstein}
Shen J, Qu Y, Zhang W, Yu Y (2018) Wasserstein distance guided representation
  learning for domain adaptation. In: Thirty-Second AAAI Conference on
  Artificial Intelligence

\bibitem[{Sugiyama and Storkey(2007)}]{sugiyama2007mixture}
Sugiyama M, Storkey AJ (2007) Mixture regression for covariate shift. In:
  Advances in Neural Information Processing Systems, pp 1337--1344

\bibitem[{Sugiyama et~al.(2008)Sugiyama, Nakajima, Kashima, Buenau, and
  Kawanabe}]{sugiyama2008direct}
Sugiyama M, Nakajima S, Kashima H, Buenau PV, Kawanabe M (2008) Direct
  importance estimation with model selection and its application to covariate
  shift adaptation. In: Advances in neural information processing systems, pp
  1433--1440

\bibitem[{Tian et~al.(2018)Tian, Ma, Zhang, and Zhan}]{tian2018deep}
Tian C, Ma J, Zhang C, Zhan P (2018) A deep neural network model for short-term
  load forecast based on long short-term memory network and convolutional
  neural network. Energies 11(12):3493

\bibitem[{Wang et~al.(2018)Wang, Liu, Bao, and Zhang}]{wang2018short}
Wang Y, Liu M, Bao Z, Zhang S (2018) Short-term load forecasting with
  multi-source data using gated recurrent unit neural networks. Energies
  11(5):1138

\bibitem[{Werbos(1990)}]{werbos1990backpropagation}
Werbos PJ (1990) Backpropagation through time: what it does and how to do it.
  Proceedings of the IEEE 78(10):1550--1560

\bibitem[{Xu et~al.(2019)Xu, Gurram, Whipps, and Chellappa}]{xu2019wasserstein}
Xu P, Gurram P, Whipps G, Chellappa R (2019) Wasserstein distance based domain
  adaptation for object detection. arXiv preprint arXiv:190908675

\bibitem[{Xu et~al.(2014)Xu, Li, Niu, and Xu}]{xu2014exploiting}
Xu Z, Li W, Niu L, Xu D (2014) Exploiting low-rank structure from latent
  domains for domain generalization. In: European Conference on Computer
  Vision, Springer, pp 628--643

\bibitem[{Yang and Hospedales(2014)}]{yang2014unified}
Yang Y, Hospedales TM (2014) A unified perspective on multi-domain and
  multi-task learning. arXiv preprint arXiv:14127489

\bibitem[{Yang and Hospedales(2016)}]{yang2016multivariate}
Yang Y, Hospedales TM (2016) Multivariate regression on the grassmannian for
  predicting novel domains. In: Proceedings of the IEEE Conference on Computer
  Vision and Pattern Recognition, pp 5071--5080

\end{thebibliography}
 
\end{document}